\documentclass[letterpaper]{article} 
\usepackage{aaai2026}  
\usepackage{times}  
\usepackage{helvet}  
\usepackage{courier}  
\usepackage[hyphens]{url}  
\usepackage{graphicx} 
\usepackage{natbib}  
\usepackage{caption} 
\usepackage{algorithm}
\usepackage{algorithmic}
\usepackage{lineno}
\usepackage{style}

\usepackage{newfloat}
\usepackage{listings}
\DeclareCaptionStyle{ruled}{labelfont=normalfont,labelsep=colon,strut=off} 
\floatstyle{ruled}
\newfloat{listing}{tb}{lst}{}
\floatname{listing}{Listing}
\title{Preference Elicitation for Step-Wise Explanations in Logic Puzzles}
\author {
    Marco Foschini\textsuperscript{\rm 1},
    Marianne Defresne\textsuperscript{\rm 2},
    Emilio Gamba\textsuperscript{\rm 3},
    Bart Bogaerts\textsuperscript{\rm 1,4},
    Tias Guns\textsuperscript{\rm 1}
}
\affiliations {
    \textsuperscript{\rm 1}KU Leuven, Dept.\ of Computer Science, Celestijnenlaan 200A, 3001 Heverlee, Belgium\\
    \textsuperscript{\rm 2}Université de Toulouse, LAAS-CNRS, Av. du Colonel Roche 7, 31400 Toulouse, France \\
    \textsuperscript{\rm 3}Flanders Make, Gaston Geenslaan 8, 3001 Heverlee, Belgium\\
    \textsuperscript{\rm 4}Vrije Universiteit Brussel, Dept.\ of Computer Science, Pleinlaan 2, Brussels, Belgium\\
    marco.foschini@kuleuven.com, 
    mdefresn@insa-toulouse.fr,
    emilio.gamba@flandersmake.be,
    bart.bogaerts@kuleuven.be,
    tias.guns@kuleuven.be
    
}

\begin{document}

\maketitle

\begin{abstract}
Step-wise explanations can explain logic puzzles and other satisfaction problems by showing how to derive decisions step by step. Each step consists of a set of constraints that derive an assignment to one or more decision variables. However, many candidate explanation steps exist, with different sets of constraints and different decisions they derive. To identify the most comprehensible one, a user-defined objective function is required to quantify the quality of each step. However, defining a good objective function is challenging. Here, interactive preference elicitation methods from the wider machine learning community can offer a way to learn user preferences from pairwise comparisons. We investigate the feasibility of this approach for step-wise explanations and address several limitations that distinguish it from elicitation for standard combinatorial problems. First, because the explanation quality is measured using multiple sub-objectives that can vary a lot in scale, we propose two dynamic normalization techniques to rescale these features and stabilize the learning process. We also observed that many generated comparisons involve similar explanations. For this reason, we introduce MACHOP (Multi-Armed CHOice Perceptron), a novel query generation strategy that integrates non-domination constraints with upper confidence bound-based diversification. We evaluate the elicitation techniques on Sudokus and Logic-Grid puzzles using artificial users, and validate them with a real-user evaluation. In both settings, MACHOP consistently produces higher-quality explanations than the standard approach.
\end{abstract}

\begin{links}
    \link{Code}{https://github.com/ML-KULeuven/MACHOP}
\end{links}

\section{Introduction}

The field of Explainable Artificial Intelligence (XAI) aims to build user trust by providing systems with explainable agency. XAI for Constraint Programming (CP)~\cite{rossi2006handbook} aims to explain why a model is unsatisfiable~\cite{DBLP:journals/jar/LiffitonS08}, why a specific variable assignment is chosen~\cite{DBLP:journals/ai/BogaertsGG21}, or why a solution is optimal~\cite{DBLP:conf/cp/BleukxDG0G23}. Without explanations, understanding these decisions can be difficult, especially given the complexity of CP models.

Step-wise explanations~\cite{DBLP:journals/ai/BogaertsGG21,BFDBG26UsingCertifyingConstraintSolversGeneratingStep-wise} can explain how decision variables, logically implied by the constraints, are derived. For instance, in nurse rostering, one explanation step might justify assigning a nurse to a night shift by noting that other nurses requested earlier shifts and the night shift must meet minimum staffing. Without this explanation, the user would have to go through the full problem specification to see what led to that decision.

Multiple explanations exist because variable assignments can be derived with different subsets of constraints~\cite{DBLP:journals/jair/GambaBG23}, though they are not all equally interpretable. Defining a criterion to assess explanation quality is crucial for generating comprehensible explanations. 
One early approach measures explanation quality by the number of used constraints~\cite{ignatiev2015smallest}. However, cardinality is not the only relevant factor, as longer explanations may contain constraints that are more familiar to the user. Later approaches~\cite{DBLP:journals/ai/BogaertsGG21,DBLP:journals/jair/GambaBG23} introduce a linear objective function to quantify human understandability, which is then optimized to identify the best explanation. 
Such function is problem-specific and requires defining and weighting each sub-objective, a complex and error-prone process~\cite{mesquita2023new}.

Interactive optimization methods~\cite{DBLP:journals/tiis/MeignanKFPG15} offer an interesting alternative, where users iteratively express preferences to find their preferred solution. We are especially interested in data-driven approaches that can learn over many related optimization problems at once. In particular, pairwise preference elicitation methods query the user to compare solution pairs, limiting the cognitive load. Such approaches have been extended to combinatorial settings through the Constructive Preference Elicitation (CPE) framework~\cite{dragone2018constructiveA}. 

We investigate the feasibility of this approach for step-wise explanations and contribute the following:
\begin{itemize}
\item Since explanation quality is measured by multiple sub-objectives that vary in scale (e.g., number of constraints, number of facts), we examine how different normalizations impact the quality of the generated explanations.
\item To generate more diverse queries, we introduce MACHOP (Multi-Armed CHOice Perceptron), a new query generation criterion for CPE based on non-domination and Upper Confidence Bound (UCB).
\item We examine the trade-off between the user's waiting time and solution quality by pre-determining the facts the solver can explain during learning. 
\item We validate our contributions through an experimental evaluation using simulated users and logic puzzles, which are standard benchmarks for explainable constraint programming~\cite{DBLP:journals/ai/BogaertsGG21,DBLP:journals/jair/GambaBG23}. A real-user evaluation is then conducted on Sudoku puzzles to demonstrate that MACHOP is able to learn human preferences.
\end{itemize}

\section{Related Work}\label{sec:related}

Preference elicitation approaches estimate preferences through user interaction. Pairwise comparisons of solutions are the most common queries, as they reduce cognitive load for the user~\cite{DBLP:conf/atal/Conitzer07}.
As common in utility theory~\cite{braziunas2007minimax}, 
user preferences are represented by a function over features, each capturing some aspect of a solution. In combinatorial settings, features are interpreted as sub-objectives, often combined with a weighted sum~\cite{DBLP:conf/sum/BenabbouL19,dragone2018constructiveB,DBLP:journals/corr/abs-2007-14778}. While non-linear functions can be learned~\cite {herin2023learning}, they are often unsupported by solvers, which is why preference elicitation for combinatorial problems relies on linear objective functions~\cite{DBLP:conf/sum/BenabbouL19,defresne2025preference}. Additionally, linear models are supported by psychological research on decision making~\cite{dawes2008robust} and remain widely used in other domains~\cite{DBLP:conf/nips/ChristianoLBMLA17,DBLP:journals/corr/abs-2403-05534}.

Learning preferences hence correspond to estimating the weight of each objective.
Approaches for multi-objective combinatorial problems include polyhedral methods~\cite{toubia2004polyhedral,DBLP:conf/sum/BenabbouL19}, which assume error-free user response, and Bayesian approaches~\cite{DBLP:journals/corr/abs-2007-14778}, which are computationally expensive. Using preference elicitation for combinatorial settings through CPE is a tractable and robust-to-noise alternative, which also enables generalization across problem instances~\cite{defresne2025preference}. 
Research on CPE has focused on improving the query generation criterion, \textit{i.e.,} the pair of solutions to show to the user, to enhance learning efficiency.~\citet{dragone2018constructiveB} generate two solutions, balancing quality and diversification.
~\citet{defresne2025preference} consider uncertainty over the solutions' utility, but it depends on a large pre-computed set of solutions.

\section{Background}
We begin by formalizing constraint satisfaction problems.
\begin{definition} 
A \textbf{constraint satisfaction problem} (CSP)~\cite{rossi2006handbook} is a triple $\CSP$ where $\variables$ is a set of decision variables, $\domains $ is a set of domains $\domainof{\variable}$ of allowed values for each variable $\variable \in \variables$, $\constraints$ is a set of constraints over a subset of $\variables$.
\end{definition}

A constraint is described by an expression (\eg $x + z \neq 1$, $a \vee b$) that restricts the values that can be assigned to its variables. 
A (partial) assignment $\assignment$ is a (partial) mapping from variables to values within their domains; each variable-value pair is called a fact. An assignment satisfies (or falsifies) a constraint if the constraint evaluates to true (or false). A solution $y$ is an assignment that satisfies all constraints. 
The set $\solset$ is the set of all such solutions. A set of constraints $\econstraints\subseteq\constraints$ is satisfiable if there exists an assignment that satisfies all constraints in $\econstraints$; otherwise it is unsatisfiable.

\subsection{Explanation steps}
\begin{figure}[t]
    \centering
    \includegraphics[width=0.40\textwidth]{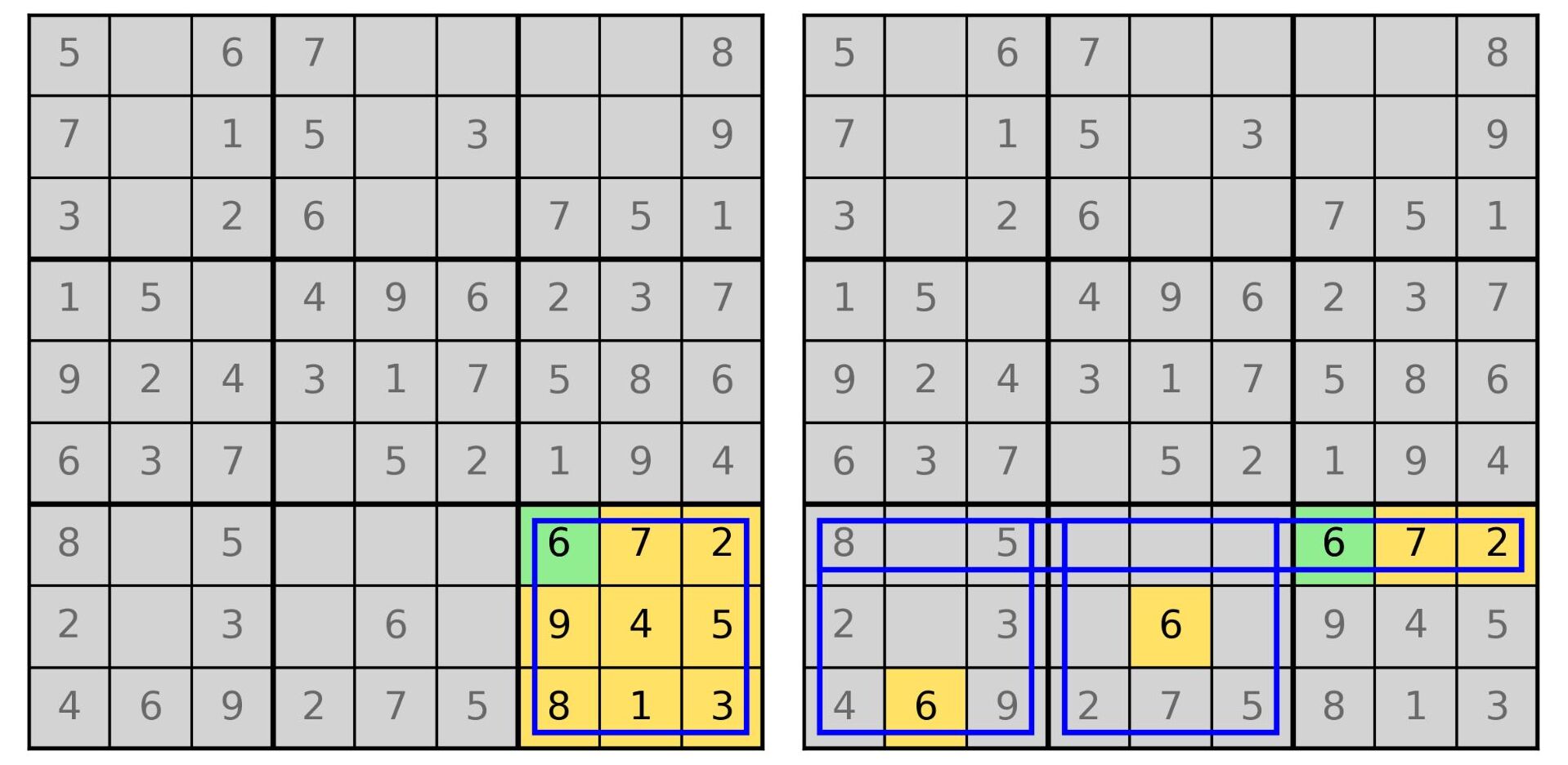}

    {\small
    \begin{tabular}{ccc}
        \hline
        \textbf{Features} & \textbf{Left} & \textbf{Right} \\ \hline
        Facts & 8 & 4 \\
        Block Cons. & 1 & 2 \\
        Row Cons. & 0 & 1 \\
        Column Cons. & 0 & 0 \\ \hline
    \end{tabular}
    }
    \caption{Two Sudoku explanation steps that explain $\mathtt{cell[7,7] = 6}$ (in green). Used facts $\efacts$ are in yellow, while used constraints $\econstraints$ are in blue. The table provides an example of the mapping from explanations to features.}
    \label{fig:explanations}
\end{figure}

Explainable facts logically follow from the CSP's constraints, \textit{i.e.} all solutions that assign the same value to the variable. These can be explained with an \textit{explanation step}:

\begin{definition}
Let $\startfacts$ be a partial assignment of a satisfiable CSP $\langle \variables, \domains, \constraints \rangle$. An \textbf{explanation step} $e$ is a triple $\explanationstep$,  also denoted $\efacts \wedge \econstraints \implies \ederived$, such that:
	\begin{itemize}
		\item
		 $\efacts \subseteq \assignment$ is a set of facts $\variable = \variablevalue$ where $\variable \in \variables$ and $\variablevalue \in \domainof{\variable} $ ;
		\item 
		$\econstraints \subseteq \constraints$ is a set of constraints.
		\item
		$\ederived$ is a set of explainable facts $x=v$, such that $\variable$ is assigned the value $\variablevalue$ in all solutions of $\efacts \cup  \econstraints$.
	\end{itemize}
\end{definition}
\noindent Fig. \ref{fig:explanations} visualizes two explanation steps of a Sudoku grid.

An explanation for \(\ederived = \{x = v\}\) can be verified through a proof by contradiction: assume \(\{x \neq v\}\), since \((\efacts \land \econstraints)\) is true, if \((\efacts\land\econstraints\land \{x\neq v\})\) is unsatisfiable, we have proven \(\{x = v\}\). Therefore, finding a (minimal) explanation step is closely related to finding \textit{Minimal Unsatisfiable Subsets}.

\begin{definition}
A subset $z \subseteq \constraints$ is a \textbf{Minimal Unsatisfiable Subset} (MUS) if $z$ is unsatisfiable and all strict subsets of $z$ are satisfiable. 
\end{definition}

Given an MUS: $\efacts\cup\econstraints\cup \{x \neq v\}$, it yields a \textit{minimal} explanation step ($\efacts \wedge \econstraints \implies \{x= v\}$). Such MUSs are the core for generating a sequence of explanation steps for a set of explainable facts $\target$~\cite{DBLP:journals/jair/GambaBG23}.

\begin{definition}
Given a CSP $\CSP$, a set of facts $\target$ to be explained, and a partial assignment $\given$, an \textbf{explanation sequence} of length $n$ is a sequence $\expseq$ of explanation steps where:
\begin{itemize}
\item $\efacts_{i} \subseteq \given \cup \bigcup_{1\leq j < i} \ederived_{j}$ for all $1 \leq i \leq n$.
\item $\bigcup_{1\leq j \leq n} \ederived_{j} = \target$.
\item the sets $\ederived_{i}$ are pairwise disjoint.
\end{itemize}
\end{definition}

Each step of the sequence derives a new assignment from $\target$, possibly using earlier derived facts ($\given \cup (\bigcup_{1\leq j < i}\ederived_{j})$).

\subsection{Preferred Explanations}
Each explainable fact in $\target$ can be explained in many ways, varying in facts and constraints used. A common approach is to compute cardinality-minimal explanations~\cite{ignatiev2015smallest}, which use the fewest facts and constraints. We will refer to these as smallest explanation steps (\textbf{SES}). Cardinality alone may not be the most relevant aspect. In Fig. \ref{fig:explanations}, the explanation on the right is smaller (4 facts + 3 constraints), but the one on the left is easier to understand (8 facts + 1 constraint). To capture different quality aspects of an explanation step, we represent it as a vector of \textit{features}.

\begin{definition}
Let $\setOES$ be the set of all explanation steps, given constraints $\constraints$, a partial assignment $\startfacts$ and a target set of facts $\target$. We define $\phi: \setOES \rightarrow \mathbb{R}^p$, mapping an explanation step to a vector of real-valued features.
\end{definition}

These features reflect dimensions of explanation quality by counting specific subsets of constraints. In Sudoku, they could represent the number of facts or the number of row/column/block constraints involved. 
These features can be used as \textit{sub-objectives} and be optimized to compute an Optimal Explanation Step (\textbf{OES}). Following the approach from~\citet{DBLP:journals/jair/GambaBG23}, preferences are expressed as a linear function $f_{w^*}$, where $w^* \in \mathbb{R}^p$ represents the importance of each sub-objective $\phi_i$.

\begin{definition}
Let $f_w(\phi(y)) = \sum_{i=1}^{p} w_i \cdot \phi_i(y)$ be a linear scalarizing function over the features of an explanation step $y \in \setOES$, $w \in \mathbb{R}_{>0}^p$. An explanation step $y \in \setOES$ is an \textbf{Optimal Explanation Step} (OES) if $\forall y' \in \setOES$ $f_w(\phi(y)) \leq f_w(\phi(y'))$.
\end{definition}

Finding an OES entails identifying the optimal explanation across all unexplained facts in $\target$. An optimal explanation step consists of selecting both an unexplained fact (\textit{e.g.,} the cell of a Sudoku) and its explanation. To compute an optimal explanation sequence, the concept of OES is extended:

\begin{definition}
An explanation sequence $\expseq^*$ is optimal according to $w$ if each triplet $(\efacts,\econstraints,\ederived) \in \expseq^*$ is an OES according to $f_w$ and across all remaining unexplained literals in $\target$.
\end{definition}

Since we treat these problems as multi-objective, the notion of domination becomes relevant.

\begin{definition}
Given two explanation steps $y^1$ and $y^2$, $\phi(y^1)$ dominates $\phi(y^2)$, if $\phi(y^2)$ is not strictly better in any of the sub-objectives~\cite{DBLP:journals/ress/KonakCS06}, \textit{i.e}:
\begin{equation}
    \begin{aligned}
        \phi(y^1) \prec \phi(y^2) : \nexists i \in [1,\ldots,p] \; \phi_i(y^2) < \phi_i(y^1)
    \end{aligned}
\end{equation}
\end{definition}

\subsection{Constructive Preference Elicitation}
Defining weights $w$ by hand is challenging and user-dependent. Instead, we aim to learn them through pairwise comparisons. The CPE framework~\cite{dragone2018constructiveA} has been proposed for learning weights for multi-objective combinatorial problems. In this context, $\phi$ maps solutions $y \in \solset$ to a vector of real-valued sub-objectives and we assume that the preferences of the user are representable as a linear function over such sub-objectives: $f_w(\phi(y)) = \sum_{i=1}^{p} w_i \cdot \phi_i(y)$. A solution $y^{+}$ is preferred over $y^{-}$, if and only if, $f_{w^*}(\phi(y^{+})) < f_{w^*}(\phi(y^{-}))$, where ${w^*}$ are the weights of the user.  

Algorithm~\ref{alg:framework} summarizes the learning process. 
Given a set of CSPs $G$, an initial estimation of weights $w^1$ (\textit{e.g.,} all equal to one) and a number of iterations $T$, the goal is to learn weights $w^T$ such that optimizing $f_{w^T}$ leads to a desirable solution. At each iteration, an instance $\solsetshort$ is selected (line 2), being a CSP $\CSP$, \textit{i.e.} with solution set $\solset$. Two solutions are then proposed to the user (line 3). They then choose their preferred solutions or indicate no preference (line 4). If the user expresses a preference, the weights are updated accordingly (line 5).

\begin{algorithm}[tb]
\caption{Constructive Preference Elicitation framework}
\label{alg:framework}
\textbf{Input}: Problems $G$, Initial weights $w^{1}$, No. of iterations $T$ \\
\textbf{Output}: Learned weights $w^{t}$
\begin{algorithmic}[1] 
\FOR{$t = 1$ to $T$}
    \STATE $\solsetshort \leftarrow  \text{Instance Selection}(G)$
    \STATE $(y^{1},y^{2}) \leftarrow \text{Query Generation}(\solsetshort,w^t)$ 
    \STATE $(y^{+},y^{-}) \leftarrow \text{Label}(y^1,y^2)$
    \STATE $w^{t+1} \leftarrow \text{Weight Update}(y^{+},y^{-},w^t)$
\ENDFOR
\STATE \textbf{return} $w^{T}$
\end{algorithmic}
\end{algorithm}

\paragraph{Query Generation.} 
In CPE, the Choice Perceptron~\cite{dragone2018constructiveB} generates a solution pair $(y_1, y_2)$ by solving the following optimization problems:
\begin{equation}
    \label{eq:cp}
    \begin{aligned}
        &y^1 = \argmin_{y \in {\solsetshort}} \; f_{w^t}(\phi(y)) \\
        &y^2 = \argmin_{y \in {\solsetshort}} \;  (1 - \gamma) \ f_{w^t}(\phi(y))- \gamma \ \alpha(y,y^1) \\   
        &\begin{array}{l}
            \text{s.t.} \quad \phi(y^2) \neq \phi(y^1) 
        \end{array} 
    \end{aligned}
\end{equation}
$y^1$ is computed by minimizing the learned objective function $f_{w^t}$, while $y^2$ is generated by minimizing $f_{w^t}$ while maximizing a diversification metric $\alpha$, namely the L1 distance:

\begin{equation}
    \label{eq:cp_L1}
    \begin{aligned}
        \alpha_{L1}(y,y^1) = \displaystyle\sum_{i = 1}^{p} |\phi_i(y^1) - \phi_i(y)|
    \end{aligned}
\end{equation}

The parameter $\gamma = \frac{1}{t}$ balances explanation quality and diversification, reducing the importance of diversification as the number of iterations increases.

\subsubsection{Weight Update.}
Given $(y^1, y^2)$, the user selects their preferred solution $y^+$, with the alternative denoted as $y^-$. To update the weights $w^t$ accordingly, the Choice Perceptron relies on an update rule inspired by the Preference Perceptron~\cite{DBLP:journals/jair/ShivaswamyJ15}. Assuming the minimization of sub-objectives, it is defined as:
\begin{equation}
\label{eq:update}
\begin{aligned}
    w^{t+1} = w^t + \eta(\phi(y^{-}) - \phi(y^{+}))
\end{aligned}
\end{equation}
where $\eta$ is the learning rate.

\section{Extending CPE to Explanation steps}
While extending CPE to explanation steps, we also propose three improvements: instance selection to reduce user wait time, enhanced diversification during query generation, and normalization of sub-objectives to stabilize learning.

\subsection{Instance Selection}
We consider a multi-instance setting, which involves both multiple CSPs (\textit{e.g.} multiple sudoku start grids) and multiple individual explanation steps of that CSP (\textit{e.g.} multiple sudoku steps). An instance $\solsetshort$ in our setting will be a CSP `state' from a CSP $g$, namely $\langle \constraints_g, \startfacts_g, \target_g \rangle$, with $\constraints_g$ its constraints, $\startfacts_g$ a partial assignment (facts) of it, and $\target_g$ the remaining explainable facts. It has a corresponding set of possible explanation steps $\setOESg$, each including one fact to explain and a subset of facts/constraints explaining it.

All instances will be assessed with the same $p$ sub-objectives. We assume stationary user preferences across instances, so a single weight vector must be learned across all.

For instance selection, we randomly choose a CSP and iteratively add one explainable fact each query, resulting in a new instance each time. Once all explainable facts are added, we pick a new random CSP.
We consider two options for determining \textit{which} explainable fact to add after each iteration:

\paragraph{Online fact selection.}
In online fact selection, we start with the initial CSP. Query generation of Algorithm~\ref{alg:framework} computes two explanation steps from all possible explanations. We store the fact with the best utility based on $w^{t+1}$, \textit{i.e.,} the updated weights after the user expresses their preference (the fact from $y^1$ is selected if $f_{w^{t+1}} \phi(y^1) < f_{w^{t+1}} \phi(y^2)$). The next returned instance will be $\setOESg$ where all stored facts for CSP $g$ are removed from $\target_g$ and added to $\startfacts_g$.

\paragraph{Offline fact selection.}
The online fact selection is computationally expensive, because the query generation on line 3 optimizes over all unexplained facts. The longer this process takes, the longer the user has to wait for the next query.

We propose two simple offline alternatives, where a sequence of facts is precomputed. Each time instance selection is called and $\setOESg$ is returned, $\target_g$ will consist of one predetermined fact, so the query generation will only need to search over the candidate explanations of this fact. The two alternatives are:

\begin{itemize}
    \item \textbf{Random}: a sequence of explainable facts is randomly selected upfront.
    \item \textbf{SES}: a sequence of facts is generated such that they are associated with the smallest explanation step. 
\end{itemize}

\subsection{Query Generation}
The default query generation is defined in Eq.~\ref{eq:cp}. We use the OCUS algorithm~\cite{DBLP:journals/jair/GambaBG23} to solve the optimal (constrained) explanation generation problems, which support linear objective functions and side constraints. To improve query generation for our setting, we propose two changes concerning how $y^2$ is computed:

\subsubsection{Non-domination criteria.}
\label{subsubsec:non-dom}
In the Choice Perceptron, $y^2$ is computed by jointly optimizing $f_{w^t}$ and $\alpha$ (Eq.~\ref{eq:cp}). The quickly decreasing $\gamma$ parameter controls the trade-off. Therefore, after a few iterations, $y^2$ often corresponds to the second-best explanation under $f_{w^t}$. We empirically observed that $y^2$ is often dominated by $y^1$, making the preference label trivial to infer. Assuming the optimization direction (minimize or maximize) for each objective is known, we can ensure non-domination between explanations with a disjunctive constraint~\cite{DBLP:journals/eor/SylvaC04}: 
\begin{equation}
     \quad \displaystyle\bigvee_{i=1}^{p} \phi_i(y^2) < \phi_i(y^1)   
\end{equation}
ensuring one of the $p$ objectives improves compared to $y^1$.

\subsubsection{Weighting schemes.}
The function $\alpha$ is used to diversify $y^2$ from $y^1$. We propose to further enhance this diversification by weighing the different components of $\alpha$ differently:
\begin{equation}
    \begin{aligned}
        &\alpha_{u}(m,y^1) = u \cdot \alpha(m,y^1) \\
    \end{aligned}
\end{equation}

This strategy uses the current estimate of the preference weights as a guide: $u = w^t$. This weighs components based on estimated importance. It favors a solution differing in the most important objectives identified so far. However, it may over-focus on objectives marked as important early on. To address this, we draw inspiration from the Upper Confidence Bound (UCB),  used in multi-armed bandits~\cite{DBLP:journals/ml/AuerCF02}. UCB focuses on high-reward actions (sub-objectives in our case) as well as those less explored. We extend it to preference elicitation by introducing MACHOP (Multi-Armed CHOice Perceptron), prioritizing the diversification of objectives that are either estimated to be important or insufficiently explored. Given $Q$, the set of all proposed queries, MACHOP is defined as:
\begin{align}
    u_{i} &= q_i + 2\sqrt{\frac{\log(|Q    |)}{N_i}} \label{eq:ucb} \\
    q_i &= \frac{\sum_{\forall(y^+,y^-) \in Q} \phi_i(y^+) < \phi_i(y^{-})}{N_i} \label{eq:pref-trade-offs} \\
    N_i &= \sum_{\forall(y^+,y^-) \in Q} \phi_i(y^+) \neq \phi_i(y^{-}) \label{eq:trade-offs}
\end{align}

Eq. \ref{eq:ucb} is the UCB formulation. In multi-armed bandits, $N_i$ is how many times the $i^{th}$ arm has been pulled, while $q_i$ is the average reward. We reinterpret $N_i$ as the count of solution pairs where the sub-objectives differ (Eq. \ref{eq:trade-offs}), indicating how many times we explore that trade-off; $q_i$ is the average number of pairs in which the solution picked has that sub-objective improved (Eq. \ref{eq:pref-trade-offs}). A higher value of $q_i$ implies that the $i^{th}$ objective is important, as explanations with improvements in that objective are more frequently selected. When $N_i = 0$, $u_i=\infty$, prompting exploration of that objective.

\subsection{Weight Update}
The weight update (Eq. ~\ref{eq:update}) is sensitive to the scale of the sub-objectives, as such $\phi$ can benefit from being \textbf{normalized} to be on the same scale. 

\paragraph{Approximate Nadir Point Normalisation.}
A standard strategy is to normalize each sub-objective by its lower ($f^{lb}$) and upper ($f^{ub}$) bound~\cite{DBLP:journals/eor/OzlenA09,defresne2025preference}.
Computing the lower bound of each sub-objective is straightforward and involves minimizing just that objective. Determining the upper bound, or nadir point, is more complex and approximations are used in practice~\cite{DBLP:journals/eor/OzlenA09}.
In our case, the sub-objectives are based on the facts and constraints, and we want to find out the maximum value for each sub-objective. For this, for each considered problem $g \in G$, we consider the set $\startfacts^{*}$ of partial assignments, where each $\startfacts_j \in \startfacts^{*}$ assigns values to all variables except one. The unassigned variable defines the corresponding target set $\target_j$. These partial assignments provide access to the maximum amount of facts; for each partial assignment $\startfacts_j \in \startfacts^{*}$, we then compute the explanation step $m$ that maximizes the objective $\phi_i$. The highest value among these is the approximate nadir point for that sub-objective.  
\begin{equation}
    \begin{aligned}
        f_{i}^{ub} &= \max_{\startfacts_j\in\startfacts^{*}} \quad \max_{m \in \mathcal{M}({\constraints,\m{I}_j,\m{T}_j})} \phi_i(m) 
    \end{aligned}
\end{equation}
This approach overestimates the upper bound, which may result in low-scale normalized $\phi_i(y)$. To avoid overestimating these bounds, we investigate two ways to normalize based on the computed pairs:

\paragraph{Cumulative Normalization.}
The upper bound of each sub-objective is defined as the maximum encountered value during all training so far. Initially, $f_i^{ub}$ is set to 1.
\begin{equation}
    \begin{aligned}
        f_{i}^{ub} &= \max(\phi_i(y^1),\phi_i(y^2),f_{i}^{ub})
    \end{aligned}
\end{equation}

\paragraph{Local Normalization.}
The upper bound of each sub-objective is defined as the maximum value of $\phi_i$ of the most recent pair $(y^1,y^2)$. If both are zero, $f_i^{ub}$ is set to 1.
\begin{equation}
    f_{i}^{ub} =
    \begin{cases}
        \max(\phi_i(y^1), \phi_i(y^2)) & \text{if } \phi_i(y^1) \neq \phi_i(y^2) \neq 0 \\
            1 & \text{otherwise}
    \end{cases}
\end{equation}

\section{Experimental Evaluation}

To experimentally evaluate our method, we address the following research questions:
\begin{itemize}
    \item [\textbf{Q1}] 
    To what extent does the non-domination constraint improve the quality of the learned explanation sequence?
    \item [\textbf{Q2}] How does the normalization affect the quality of the learned explanation sequence?
    \item [\textbf{Q3}] How do the weighting schemes affect the quality of the learned explanation sequence? 
    \item [\textbf{Q4}] What trade-off between runtime and explanation quality can be reached by offline fact selection?
    \item  [\textbf{Q5}] How does our method perform when learning the preferences of real users?
\end{itemize}

\subsection{Problems}
\paragraph{Sudoku.}
We generate Sudoku puzzles with QQWing~\cite{ostermillerqqwing}, and produce explanation steps by using a Boolean encoding of the problem.

\paragraph{Logic-Grid puzzles.} A Logic-Grid puzzle, also known as Einstein puzzles, consists of sentences (\textit{clues}) over a set of occurring entities. The goal is to determine the associations between these entities. Constraints can be classified into three categories~\cite{DBLP:journals/ai/BogaertsGG21}: \textit{clues}, \textit{transitivity constraints} and \textit{bijectivity constraints}. A fact is \textit{positive} if one entity is associated with another, or \textit{negative} otherwise. We consider the problems from~\citet{DBLP:journals/jair/GambaBG23}. An example is in the appendix.

\subsection{Definition of the sub-objectives}\label{subsec:obj def}
To define explanation sub-objectives, we measure the distance from any constraint to the explained fact $x=v$. Constraints at distance one are those that involve variable $x$. We consider the (bipartite) variable-constraint graph where variables are linked to the constraints they appear in, allowing us to group constraints by their distance in this graph to $x$. For facts, \ie $y=z$, we group the facts based on the constraint-graph distance of $x$ to $y$ as well as on their values $v$ and $z$.

\paragraph{Sub-objectives for Sudoku.}
Constraints for Sudoku are either \texttt{facts} or $\texttt{alldifferent}$ constraints. We group the constraints into \textit{adjacent} constraints (constraints at graph distance 1 from the fact to explain $x=v$) and other constraints (distance $>1$). Facts are categorized based on their adjacency to the explained fact and whether they assign the same value as the one being explained. Constraints are split into separate subgroups for row, column and block constraints.

\paragraph{Sub-objectives for Logic-Grid puzzles.}\label{sec:Sudoku-feature-explanations}
Following the approach used for Sudoku, we categorize constraints and facts by distance in the constraint graph. Facts (Boolean variables in this case) are further classified as either being True or False. Constraints are split into: transitivity, bijectivity and clues.

The full list of features is provided in the appendix.

\subsection{User simulation}
For simulated users, the response is modelled by an oracle based on the Bradley-Terry model~\cite{bradley1952rank} with indifference. Given a query $(y^1,y^2)$, the probability of a user being indifferent is defined as~\cite{DBLP:journals/jmlr/GuoS10}:
\begin{equation}
P(\phi(y^1) \sim \phi(y^2)) = e^{-\beta |f_{w^*}(\phi(y^2)) - f_{w^*}(\phi(y^1))|}
\label{eq:bradley}
\end{equation}
with $w^*$ as the oracle’s true preference weights. The probability of indifference depends on the difference in explanation utility according to $f_{w^*}$. We set $\beta=1$. We follow a similar approach to~\citet{DBLP:conf/nips/ChristianoLBMLA17}, assuming a 10\% chance of mislabeling. Conceptually, real users make errors constantly, regardless of sub-objective distance. 

To represent clear preferences for some sub-objectives, weights $w^*$ are generated with an exponential distribution: each component $w^*_i$ is $10^{j}$, where $j$ is chosen uniformly randomly between -2 and 2~\cite{DBLP:conf/icaps/MischekM24}.

\begin{figure*}[htb]
    \begin{subfigure}[t]{0.5\textwidth}
        \centering
        \includegraphics[width=0.8\textwidth]{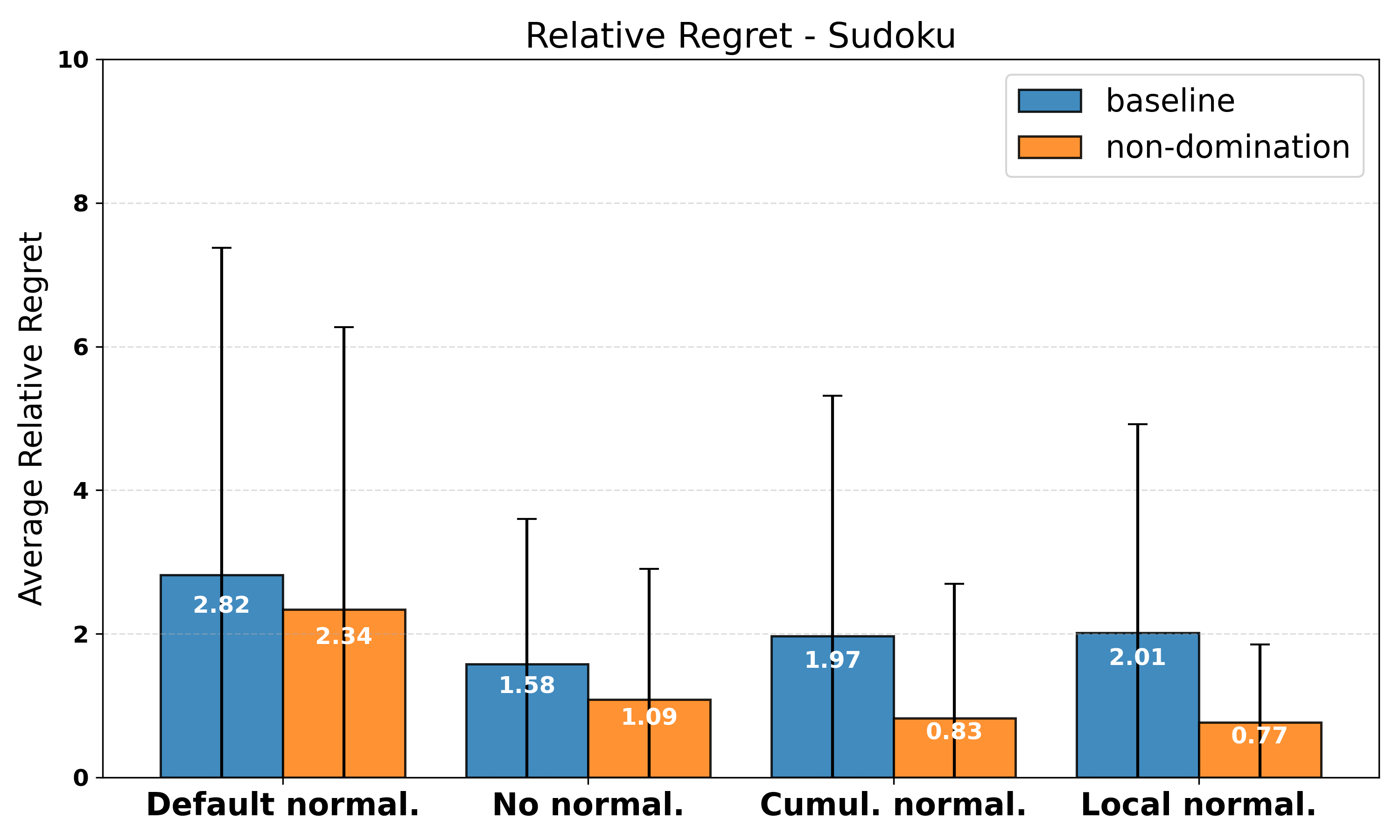}
        \label{fig:RQ2_sudoku}
    \end{subfigure}
    \begin{subfigure}[t]{0.5\textwidth}
        \centering
        \includegraphics[width=0.8\textwidth]{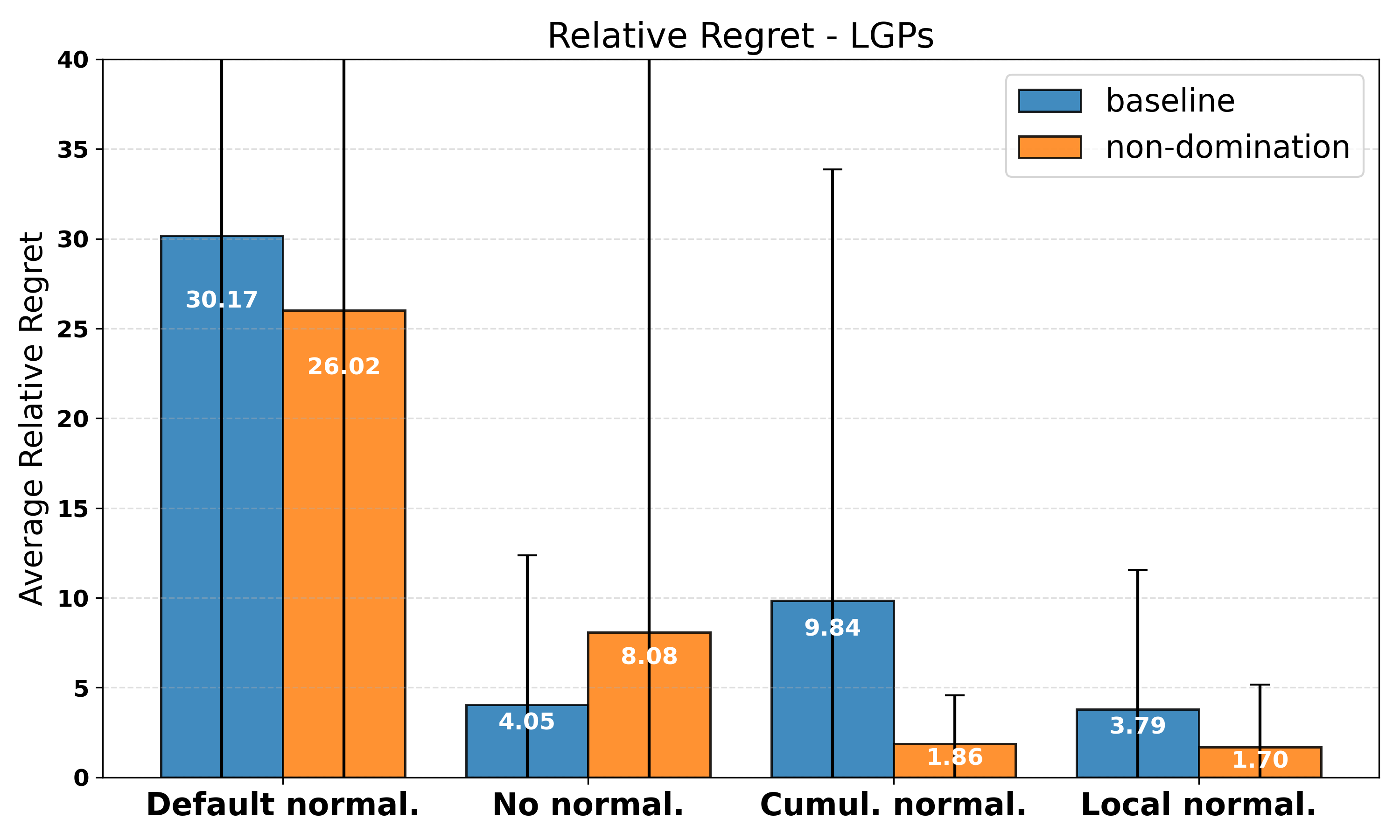}
        \label{fig:RQ2_LGP}
    \end{subfigure}
    \caption{Average relative regret for explanation sequences.}
    \label{fig:RQ2}
\end{figure*}

\subsection{Metric}
The quality of an explanation, generated with the learned objective function $f_{w}$, is assessed with relative regret:
\begin{equation}
    \begin{aligned}
        \textrm{R}(\insetOES,w^{*},w) &= \frac{f_{w^{*}}(\phi(y)) - f_{w^{*}}(\phi(y^{*}))}{f_{w^{*}}(\phi(y^{*}))} \\
        y &= \argmin_{m \in \setOES} f_w(\phi(m)) \\
        y^{*} &= \argmin_{m \in \setOES} f_{w^*}(\phi(m)) \\
    \end{aligned}
\end{equation}

To evaluate the learned objective function in generating a desirable explanation sequence, we compare it to the explanation sequence that $w^{*}$ would generate. Given a starting instance $\triplesOES$ of a given CSP, and explanation sequence $E^{*} = \expseqstart$ that is optimal according to $w^{*}$, we compute the average (sequential) relative regret as:

\begin{equation}
    \begin{aligned}
        &\textrm{R\_seq}(\insetOES, E^{*}, w^{*}, w) = \\
        &\begin{array}{l}
    \frac{1}{n}\sum\limits_{i=1}^{n} \textrm{R} \left( \constraints, \startfacts \bigcup\limits_{1 \leq j < i} \ederived^*_j,  \target - \bigcup\limits_{1 \leq j < i} \ederived^*_j, w^*, w \right)
\end{array} 
    \end{aligned}
\end{equation}

To generate a new explanation step, we add all facts already explained by $E^{*}$ ($\startfacts \cup_{1 \leq j < i} \ederived^*_j$) as facts and generate an explanation over the unexplained ones ($\target - \bigcup\limits_{1 \leq j < i} \ederived^*_j$).

\subsection{Results}

\paragraph{Experimental Setup.}
Experiments were run on systems with Intel(R) Xeon(R) Silver 4514Y and 256 GB of memory. All methods were implemented in Python, using CPMpy 0.9.24~\cite{guns2019cpmpy}. Split OCUS~\cite{DBLP:journals/jair/GambaBG23} used Exact~\cite{exact_repo} for the SAT calls and GurobiPy~\cite{gurobi} for the MIP calls. Each training loop consists of 100 queries. A hyperparameter search over $\eta \in [0.1,0.5,1,5,10]$ was performed by considering 4 oracles and 3 runs. Final results are obtained by aggregating over 10 oracles and 5 runs.

\paragraph{Q1 (non-domination criteria).}\label{sec:RQ1}
We assess the impact of preventing dominated explanations using a disjunctive constraint, by comparing the quality of explanations learned with the \textbf{baseline} and its \textbf{non-domination} variant. The analysis is conducted across all normalization strategies. As shown in Fig.~\ref{fig:RQ2}, adding the disjunctive constraints yields a significantly lower regret in $7$ out of $8$ setups. These results support its use, which will be applied in the subsequent experiments. 

\begin{figure*}[htb]
    \centering
    \begin{minipage}[t]{0.38\textwidth}
        \vspace{0pt}
        \centering
        \includegraphics[width=\textwidth]{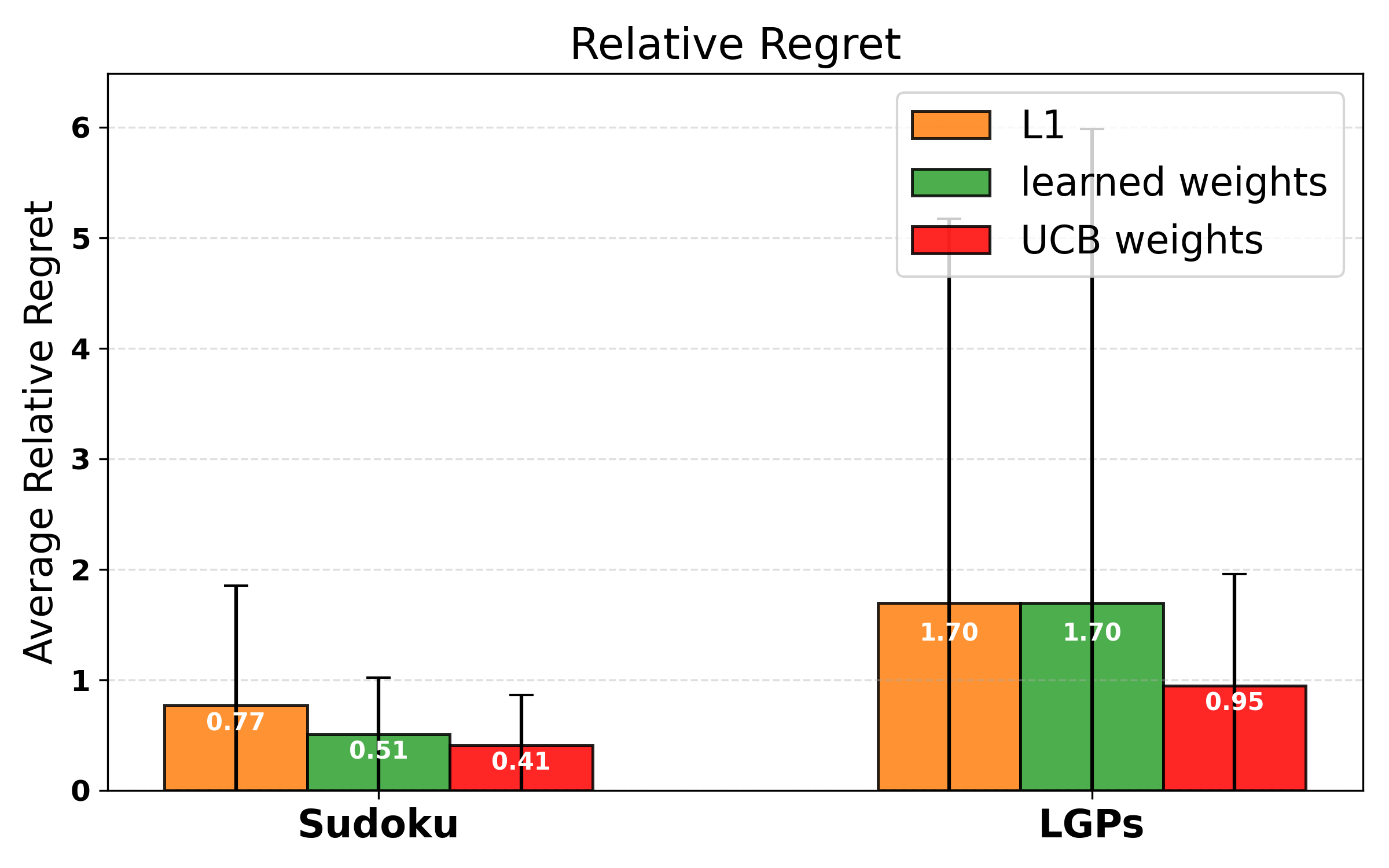}
        \caption{Relative regret for the different diversification strategies for Sudoku and LGPs.}
        \label{fig:RQ3}
    \end{minipage}
    \hfill
    \begin{minipage}[t]{0.58\textwidth}
        \vspace{0pt} 
        \centering

        \renewcommand{\arraystretch}{0.9}
        \setlength{\tabcolsep}{4pt}
        \captionof{table}{Relative regret for Sudoku and LGPs.}
        \label{tab:RQ3t}
        \begin{tabular}{@{}lccccc@{}}
            \toprule
            & \multicolumn{2}{c}{\textbf{Sudoku}} & & \multicolumn{2}{c}{\textbf{LGPs}} \\
            \cmidrule{2-3} \cmidrule{5-6}
            & Regret & & & Regret & \\
            \midrule
            Choice Perceptron & $2.0 \pm 2.9$ & & & $3.8 \pm 7.7$ & \\
            \textbf{MACHOP} & $\mathbf{0.4 \pm 0.4}$ & & & $\mathbf{0.9 \pm 1.0}$ & \\
            \bottomrule
        \end{tabular}

        \vspace{0.8em}

        \renewcommand{\arraystretch}{0.9}
        \setlength{\tabcolsep}{4pt}
        \captionof{table}{Relative regret and average query generation time for MACHOP}
        \label{tab:RQ4}
        \begin{tabular}{@{}lccccc@{}}
        \toprule
        & \multicolumn{2}{c}{\textbf{Sudoku}} & & \multicolumn{2}{c}{\textbf{LGPs}} \\
        \cmidrule{2-3} \cmidrule{5-6}
        \textbf{Chosen Fact} & Time (s) & Regret & & Time (s) & Regret \\
        \midrule
        Online  & $35.6 \pm 38.3$ & $\mathbf{0.4 \pm 0.4}$ & & $49.2 \pm 15.1$ & $\mathbf{0.9 \pm 1.0}$ \\
        Offline - Random    & $44.2 \pm 44.7$ & $0.7 \pm 0.8$ & & $11.2 \pm 1.9$ & $2.7 \pm 9.7$ \\
        Offline - SES       & $\mathbf{12.5 \pm 26.7}$ & $0.6 \pm 0.4$ & & $\mathbf{8.3 \pm 1.0}$ & $1.4 \pm 3.2$ \\
        \bottomrule
        \end{tabular}

    \end{minipage}
\end{figure*}

\begin{table*}[bt!]
\centering
\caption{User study: \% picked solutions from SES with MACHOP and Choice Perceptron (ChoPerc) and generation time.}
\label{tab:RQ5}
\renewcommand{\arraystretch}{0.9} 
\setlength{\tabcolsep}{4pt}      
\begin{tabular}{@{}c ccc ccc cc@{}}
\toprule
& \multicolumn{3}{c}{\textbf{Evaluation: SES vs MACHOP}} & \multicolumn{3}{c}{\textbf{Evaluation: SES vs ChoPerc}} & \multicolumn{2}{c}{\textbf{Query Generation Time}} \\
\cmidrule(lr){2-4} \cmidrule(lr){5-7} \cmidrule(lr){8-9}
\textbf{Query} & SES (\%) & MACHOP (\%) & Indiff. (\%) & SES (\%) & ChoPerc (\%) & Indiff. (\%) & MACHOP (s) & ChoPerc (s) \\
\midrule
10 & \textbf{52.2 ± 19.8} & 25.2 ± 16.6 & 22.6 ± 12.5 & \textbf{63.3 ± 18.2} & 16.2 ± 14.8 & 20.5 ± 12.0 & 1.4 ± 0.5 & 0.9 ± 0.3 \\
30 & 10.8 ± 12.2 & \textbf{70.7 ± 18.5} & 18.5 ± 13.8 & 33.2 ± 17.1 & \textbf{44.8 ± 21.2} & 22.1 ± 14.1 & 2.6 ± 0.9 & 1.4 ± 0.6 \\
50 & 13.2 ± 11.2 & \textbf{72.6 ± 14.9} & 14.3 ± 9.3 & \textbf{42.5 ± 25.2} & 38.8 ± 25.7 & 18.7 ± 13.0 & 3.0 ± 1.0 & 1.3 ± 0.5 \\
\bottomrule
\end{tabular}
\end{table*}

\paragraph{Q2 (normalization).} We evaluate the considered normalization strategies. We compare the \textbf{default} normalization strategy based on (approximate) nadir points with \textbf{no normalization} and our two proposed normalizations, \textbf{cumulative} and \textbf{local}. Results in Fig.~\ref{fig:RQ2} show that the default normalization performs poorly for our setting, with significantly higher regret. This effect is especially pronounced in Logic-Grid puzzles, where feature value ranges are broader. Similarly, \textbf{no normalization} ignores feature scale, which negatively impacts the performance of the non-domination variant. In contrast, both proposed strategies yield the lowest regret, with \textbf{local normalization} performing best on average. We therefore adopt it in the following sections.

\paragraph{Q3 (weighting schemes).} We compare three weighting schemes for the diversification metric: \textbf{no weights}, \textbf{learned weights} and \textbf{UCB weights}. Results are shown in Fig.~\ref{fig:RQ3}. Not using any acquired knowledge to guide the diversification (no weights) results in the highest regret for both Sudoku and LGP. Using the learned weights is beneficial for Sudoku but not for LGPs, while UCB weights consistently reduce regret by about $40\%$ for both problems. This suggests that guiding diversification towards both important and unexplored objectives is effective. Overall, MACHOP (non-domination with local normalisation and UCB weighting scheme) reduces the regret by 80\% compared to the Choice Perceptron for both Sudoku and LGPs, as shown in Table~\ref{tab:RQ3t}. 

\paragraph{Q4 (offline fact selection).}

We report MACHOP's query generation time in Table~\ref{tab:RQ4}. When allowing \textbf{online} selection of the fact to explain, the waiting time can exceed one minute. Fixing the fact to explain is effective, as both offline sequences (\textbf{random} and \textbf{SES}) cut query generation time for Logic-Grid puzzles. For Sudoku, only SES speeds up generation time. We observed that random facts can require complex explanations that are costly to compute. However, both offline sequences experience a slight increase in regret compared to the online fact selection. When comparing with results from Fig.\ref{fig:RQ3}, MACHOP + SES ranks second for LGPs and third for Sudoku, meaning that it offers a reasonable speed/quality trade-off.

\paragraph{Q5 (user evaluation).}
We validate the practical relevance of our approach through a study with 30 participants on Sudoku. Preferences were elicited interactively using both Choice Perceptron and MACHOP, with up to 50 queries. For evaluation, we use SES as a reference point to assess how well the learned objective functions of MACHOP and Choice Perceptron align with user preferences, as SES explanations are reasonable baselines in these domains~\cite{DBLP:journals/jair/GambaBG23}. 

We generate an explanation sequence using SES for a new Sudoku puzzle, comprising 56 steps. For each step of this sequence, we extract $\triplesOES$ and generate explanations using each learned objective function. Users are then presented with the learned explanation and SES and asked to pick one. This evaluation was conducted after 10, 30 and 50 pairs as depicted in Table~\ref{tab:RQ5}. Including both training and evaluation, each user labelled around 400 pairs, taking 45-90 minutes.

As Table~\ref{tab:RQ5} shows, when evaluating after $10$ queries, SES explanations were preferred more often. After $30$, both MACHOP and Choice Perceptron aligned better with users' preferences, with learned explanations being preferred more often. MACHOP achieved stronger alignment than Choice Perceptron, with its explanations preferred over SES in 70.7\% of cases, compared to 44.8\% of the Choice Perceptron. At 50 queries, MACHOP's performance stayed stable, while Choice Perceptron appears to overfit, likely due to excessive exploitation.  For both 30 and 50 queries, no user selected explanations from the Choice Perceptron more frequently than from MACHOP. Statistical tests support this: one-sided Wilcoxon signed-rank~\cite{wilcoxon1992individual} gives $p < 10^{-3}$ at 10 queries, $p < 10^{-6}$ at 30 and 50 queries. Cliff’s delta~\cite{cliff1993dominance} is positive in all cases (0.37, 0.62, 0.71).

We also observe that query generation time is low, up to $10$ times smaller compared to the oracle experiments. This is due to the artificial oracles' generated utility function sometimes preferring complex explanations, which are costly to compute. With the runtimes observed in our user experiments, real-time learning of preferences becomes feasible.

\section{Conclusion}
Generating human-understandable step-wise explanations is challenging. Existing methods rely on either cardinality-minimal explanations or problem-specific objectives.
We address this by adapting the Constructive Preference Elicitation framework for step-wise explanations. However, the wide range of the sub-objectives can hinder learning, motivating new normalization strategies. Additionally, the state-of-the-art CPE method frequently suggests overly similar explanations; we propose a new query generation strategy based on non-domination and UCB.
Experimental results show that our contributions lead to higher-quality explanations in both synthetic and real-user evaluations.

Future work can explore learning preferences as a non-linear utility function, which tends to be more computationally expensive to optimize. 
Exploring how to define such a function can further help capture aspects missed by the given sub-objectives. 
Perhaps learning could be further sped up by actively choosing which instance to generate a query for next.
Additionally, using queries where users express no preference could accelerate learning too. Finally, the real-user evaluation paves the way for applying MACHOP to more practical and larger-scale scenarios, such as explanations for industrial problems.

\section*{Acknowledgments}
This research received funding from Flemish Goverment under ``Onderzoeksprogramma Artificiële Intellgentie (AI) Vlaanderen.'', from Fonds Wetenschappelijk Onderzoek -- Vlaanderen (projects G064925N and G070521N) and from the European Union (ERC, CertiFOX, 101122653; ERC, CHAT-Opt, 01002802). Views and opinions expressed are however those of the author(s) only and do not necessarily reflect those of the European Union or the European Research Council. Neither the European Union nor the granting authority can be held responsible for them.

We also thank the thirty labellers for their time, which made this publication possible.

\bibliography{ref}

\clearpage
\setcounter{secnumdepth}{2}
\title{Appendix}
\section{Appendix}
\subsection{Example of Logic grid puzzle}
Given the following clues:
\begin{itemize}
    \item  The home visited in April was either Markmanor or the home haunted by Brunhilde.
    \item The home on Circle Drive was investigated sometime before Wolfenden.
    \item Of the building haunted by Lady Grey and the building haunted by Victor, one was Markmanor and the other was visited in January.
    \item The house haunted by Victor was visited 1 month after the house haunted by Lady Grey.
    \item Of the home on Bird Road and Barnhill, one was visited in January and the other was haunted by Brunhilde.
    \item Markmanor was visited 1 month after the home on Grant Place.
    \item The house visited in march wasn't located on Circle Drive.
    \item The building visited in May wasn't located on Fifth Avenue.
    \item Hughenden wasn't investigated in march.
    \item Wolfenden was haunted by Brunhilde.
    \item Hughenden wasn't haunted by Abigail.
\end{itemize}
The goal is to find:
\begin{itemize}
    \item When a house was visited.
    \item In which street is a house located.
    \item Which ghost haunts which house.
\end{itemize}
\newpage
\subsection{Sub-objectives for Sudoku}
\begin{table}[htbp]
\centering
\label{table:Sudoku_features}
{\small
\begin{tabular}{p{0.38\columnwidth} p{0.50\columnwidth}}
\toprule
\textbf{Feature name} & \textbf{Description} \\
\midrule
Adj. Facts Other Value & Adjacent facts with different value from the explained one \\
Other Facts Same Value & Non-adjacent facts with the same value as the explained one \\
Other Facts Other Value & Non-adjacent facts with different value from the explained one \\
Adj. Block Cons & Adjacent block constraints \\
Adj. Row Cons & Adjacent row constraints \\
Adj. Col Cons & Adjacent column constraints \\
Other Block Cons & Non-adjacent block constraints \\
Other Row Cons & Non-adjacent row constraints \\
Other Col Cons & Non-adjacent column constraints \\
Adj. Facts From Block  & Adjacent facts from blocks \\
Adj. Facts From Row  & Adjacent facts from rows \\
Adj. Facts From Col. & Adjacent facts from columns \\
\bottomrule
\end{tabular}
}
\end{table}

\subsection{Sub-objectives for logic-grid puzzles.}

\begin{table}[htbp]
\centering
\label{table:lgps_features}
{\small
\begin{tabular}{p{0.38\columnwidth} p{0.50\columnwidth}}
\toprule
\textbf{Feature name} & \textbf{Description} \\
\midrule
Adj. Negative Facts & adjacent negative facts \\
Other Positive Facts & non-adjacent positive facts \\
Other Negative Facts & non-adjacent negative facts \\
Adj. Bijectivity & adjacent bijectivity constraints \\
Adj. Transitivity & adjacent transitivity constraints \\
Adj. Clues & adjacent clues constraints \\
Other Bijectivity & non-adjacent bijectivity constraints \\
Other Transitivity & non-adjacent transitivity constraints \\
Other Clues & non-adjacent clues constraints \\
Adj. Facts From Bijectivity  & adjacent facts from bijectivity constraints \\
Adj. Facts From Transitivity & adjacent facts from transitivity constraints \\
Adj. Facts From Clues & adjacent facts from clues \\
\bottomrule
\end{tabular}
}
\end{table}

\newpage
\subsection{Detailed Hyperparameter List - Sudoku}
\begin{table}[!htbp]
\centering
\begin{tabular}{l l c}
\hline
\textbf{Normalization} & \textbf{Method} & \textbf{Learning Rate} \\
\hline
\multirow{2}{*}{Default Norm.} & Choice Perceptron        & 0.1 \\
                        & Non-domination   & 0.5 \\
\hline
\multirow{2}{*}{No Norm.} & Choice Perceptron        &  0.1\\
                        & Non-domination   & 0.1 \\
\hline
\multirow{2}{*}{Cumulative Norm.} & Choice Perceptron        & 0.5 \\
                        & Non-domination   &  0.5 \\
\hline
\multirow{6}{*}{Local Norm.} & Choice Perceptron        & 0.5 \\
                        & Non-domination   & 10 \\
                        & Learned weights & 5 \\
                        & UCB weights &  0.5 \\
                        & Random - UCB weights &  10 \\
                        & SES - UCB weights &  10 \\
                        
\hline
\end{tabular}
\label{tab:normalization_methods_sudoku}
\end{table}

\subsection{Detailed Hyperparameter List - LGPs}
\begin{table}[h]
\centering
\begin{tabular}{l l c}
\hline
\textbf{Normalization} & \textbf{Method} & \textbf{Learning Rate} \\
\hline
\multirow{2}{*}{Default Norm.} & Choice Perceptron        & 0.1 \\
                        & Non-domination   & 0.5 \\
\hline
\multirow{2}{*}{No Norm.} & Choice Perceptron        &  0.1\\
                        & Non-domination   & 0.1 \\
\hline
\multirow{2}{*}{Cumulative Norm.} & Choice Perceptron        & 10 \\
                        & Non-domination   &  10 \\

\hline
\multirow{6}{*}{Local Norm.} & Choice Perceptron        & 0.1 \\
                        & Non-domination   & 5 \\
                        & Learned weights & 0.1 \\
                        & UCB weights &  0.5 \\
                        & Random - UCB weights &  0.5 \\
                        & SES - UCB weights &  0.5 \\
\hline
\end{tabular}
\label{tab:normalization_methods}
\end{table}

\newpage
\subsection{Examples of proposed pairs}
\begin{figure}[h!]
\centering
\begin{subfigure}{0.5\textwidth}
    \centering
    \includegraphics[width=\linewidth]{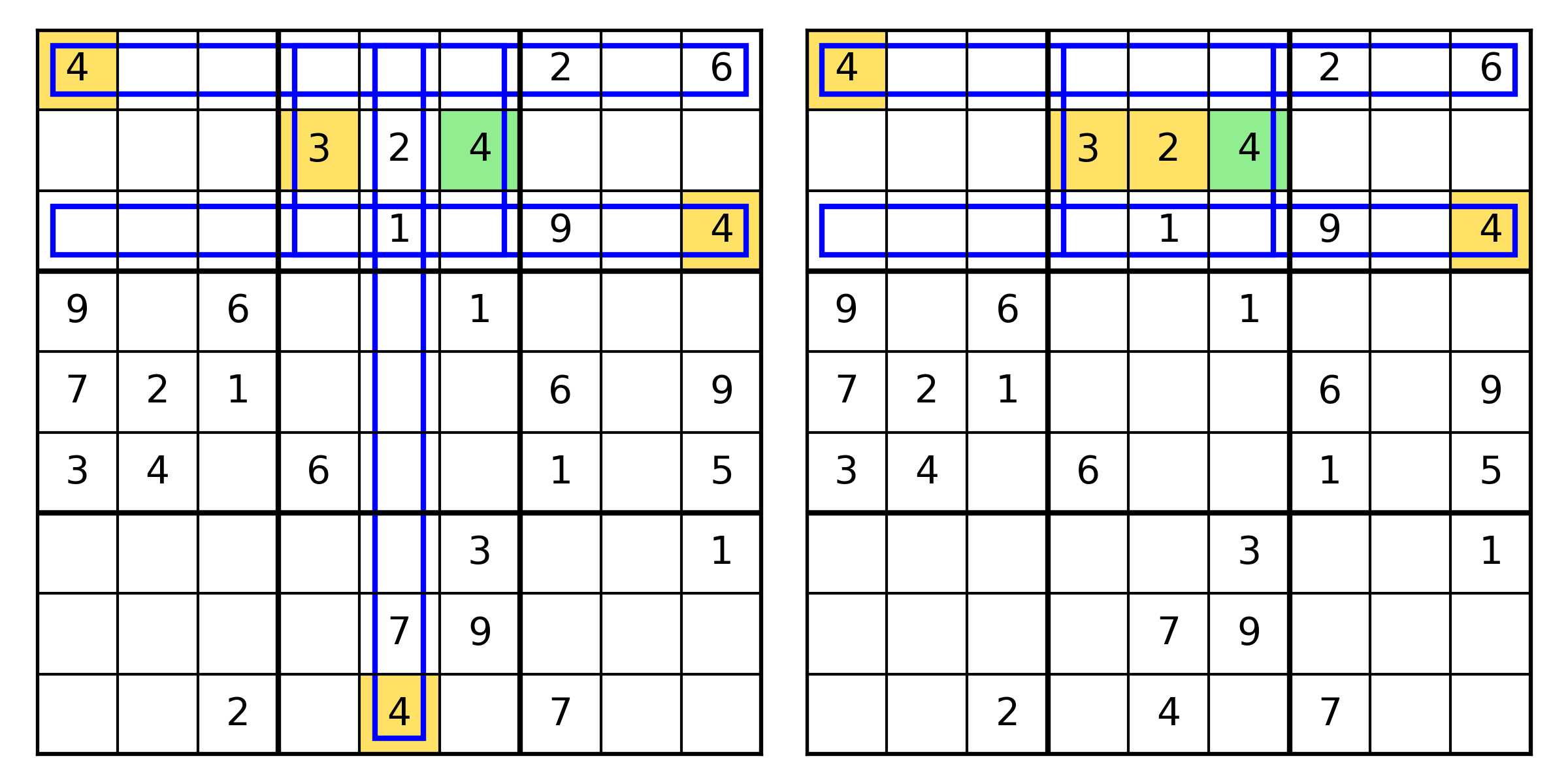}
    \label{fig:htree1}
\end{subfigure}%
\hspace{0.03\textwidth}
\begin{subfigure}{0.5\textwidth}
    \centering
    \includegraphics[width=\linewidth]{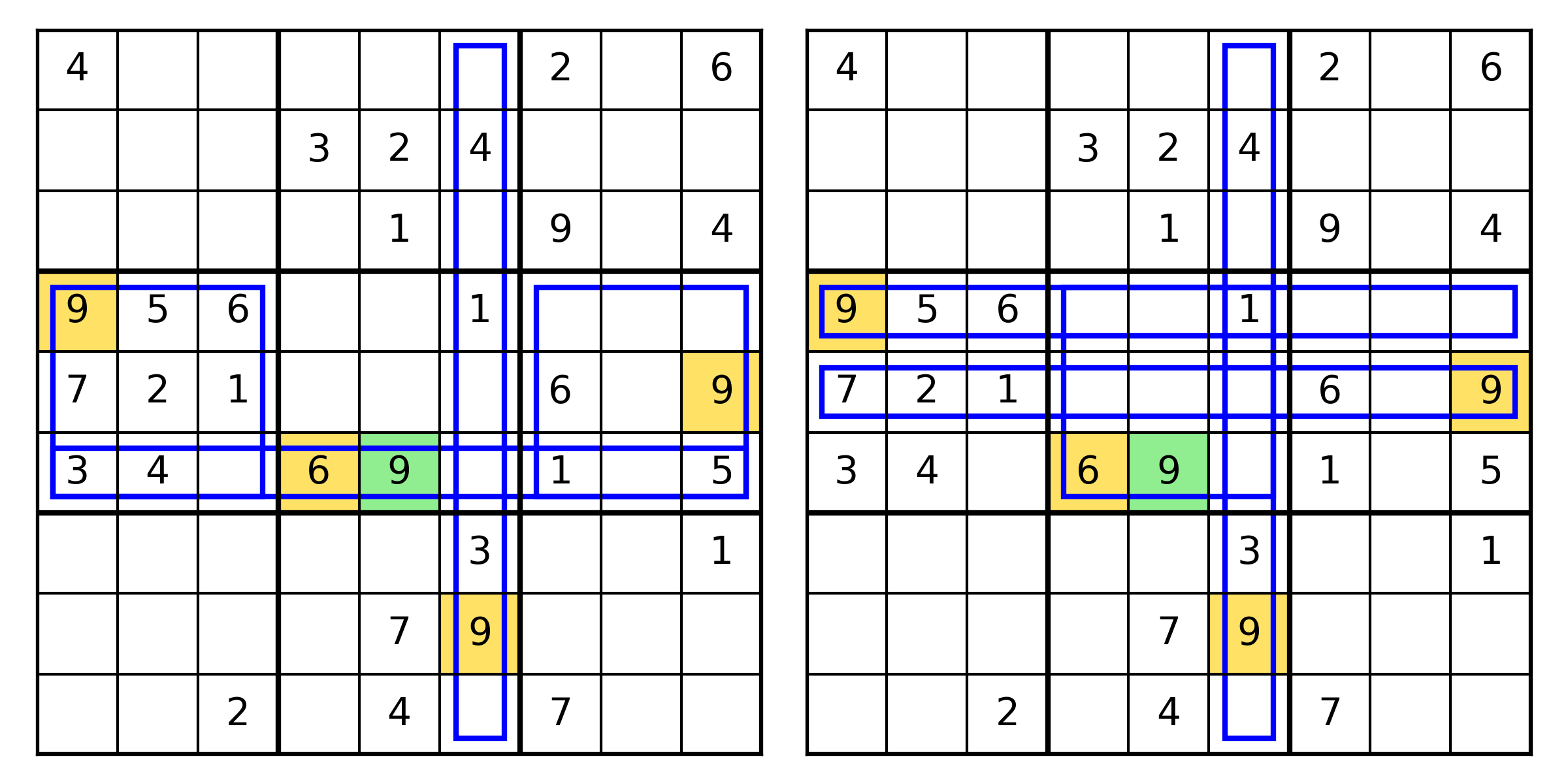}
    \label{fig:htree2}
\end{subfigure}%
\hspace{0.03\textwidth}
\begin{subfigure}{0.5\textwidth}
    \centering
    \includegraphics[width=\linewidth]{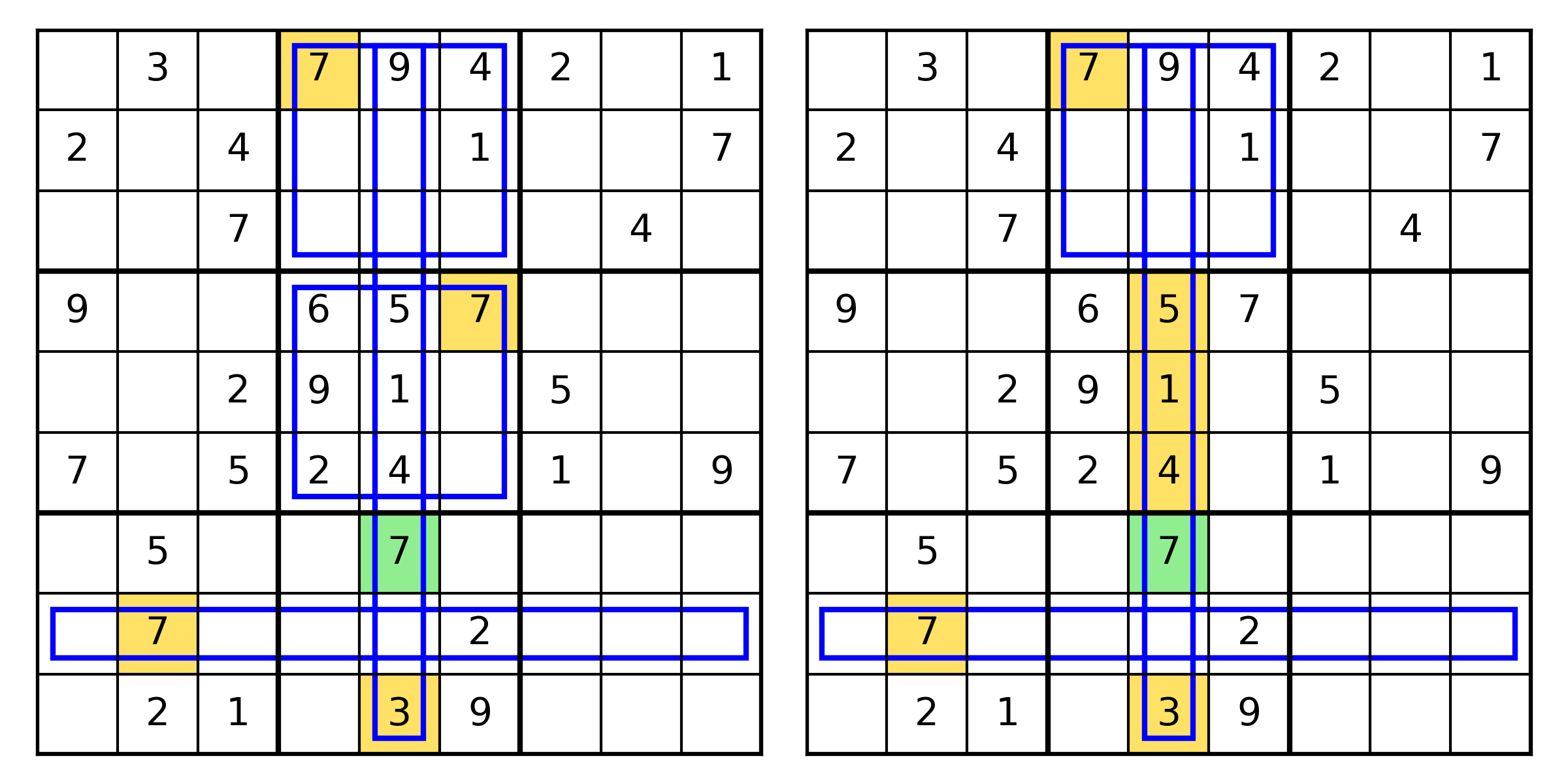}
    \label{fig:htree3}
\end{subfigure}
\label{fig:htree_all}
\end{figure}

\begin{figure*}[t]
    \centering
    \includegraphics[width=0.40\textwidth]{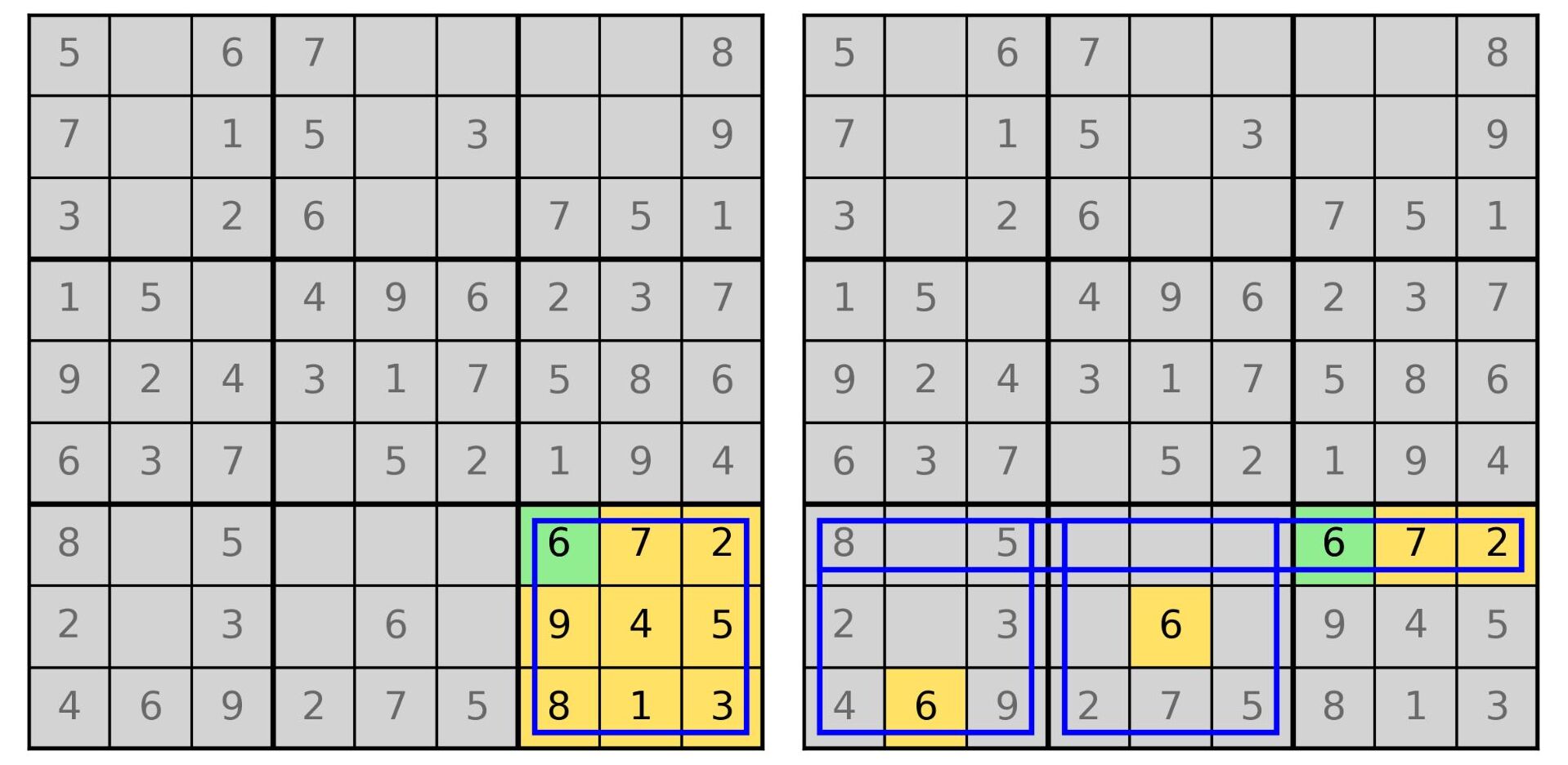}

    \begin{tabular}{ccc}
        \hline
        \textbf{Features} & \textbf{Left} & \textbf{Right} \\ \hline
        Adj. Facts Other Value & 8 & 2 \\
        Other Facts Same Value & 0 & 2 \\
        Other Facts Other Value & 0 & 0 \\
        Adj. Block Cons. & 1 & 0 \\
        Adj. Row Cons. & 0 & 1 \\
        Adj. Col Cons. & 0 & 0 \\
        Other Block Cons. & 0 & 2 \\
        Other Row Cons. & 0 & 0 \\
        Other Col Cons. & 0 & 0 \\
        Adj. facts from row & 2 & 2 \\
        Adj. facts from block & 8 & 2 \\
        Adj. facts from colum & 2 & 0 \\ 
        \hline
    \end{tabular}
    \caption{Two Sudoku explanation steps that explain $\mathtt{cell[7,7] = 6}$ (in green). Used facts $\efacts$ are in yellow, while used constraints $\econstraints$ are in blue. The table provides the exact mapping used from explanations to features.}
    \label{fig:explanations_appendix}
\end{figure*}

\begin{table*}[bt!]
\centering
\caption{Oracle evaluation: regret summary across users for Sudoku and Logic Grid Puzzles.}
\label{tab:regret_oracle_combined}
\renewcommand{\arraystretch}{0.9}
\setlength{\tabcolsep}{4pt}
\begin{tabular}{@{}l c c c c c c c c c c c c c@{}}
\toprule
& \multicolumn{6}{c}{\textbf{Sudoku}} & & \multicolumn{6}{c}{\textbf{Logic Grid Puzzles}} \\
\cmidrule(lr){2-7} \cmidrule(lr){9-14}
& \textbf{Mean} & \textbf{Median} & \multicolumn{2}{c}{\textbf{Percentile}} & \multicolumn{2}{c}{\textbf{Range}} & & 
\textbf{Mean} & \textbf{Median} & \multicolumn{2}{c}{\textbf{Percentile}} & \multicolumn{2}{c}{\textbf{Range}} \\
\cmidrule(lr){4-5} \cmidrule(lr){6-7} \cmidrule(lr){11-12} \cmidrule(lr){13-14}
\textbf{Methods} & & & \textbf{25th} & \textbf{75th} & \textbf{Min} & \textbf{Max} & &
& & \textbf{25th} & \textbf{75th} & \textbf{Min} & \textbf{Max} \\
\midrule
Choice Perceptron Online & 2.013 & 1.019 & 0.432 & 1.869 & 0.084 & 13.902 & &
3.795 & 0.880 & 0.313 & 1.283 & 0.087 & 27.408 \\
Learned Weights Online & 0.506 & 0.321 & 0.156 & 0.682 & 0.006 & 2.705 & &
1.696 & 0.507 & 0.161 & 1.162 & 0.079 & 25.664 \\
MACHOP Online & 0.410 & 0.266 & 0.146 & 0.480 & 0.007 & 2.663 & &
0.947 & 0.578 & 0.243 & 1.182 & 0.059 & 3.821 \\
MACHOP Offline -- SES & 0.588 & 0.543 & 0.278 & 0.796 & 0.065 & 1.833 & &
1.359 & 0.475 & 0.154 & 1.110 & 0.044 & 16.466 \\
\bottomrule
\end{tabular}
\end{table*}

\begin{table*}[bt!]
\centering
\caption{Real-user evaluation: summary regarding the percentage of picked solutions proposed by MACHOP and Choice Perceptron}
\label{tab:machop_vs_baseline}
\renewcommand{\arraystretch}{0.9}
\setlength{\tabcolsep}{5pt}

\begin{minipage}{0.48\textwidth}
\centering
\textbf{MACHOP}\\[3pt]
\begin{tabular}{@{}c cccccc@{}}
\toprule
\textbf{Query} & \textbf{Mean} & \textbf{Median} &
\multicolumn{2}{c}{\textbf{Percentile}} &
\multicolumn{2}{c}{\textbf{Range}} \\
\cmidrule(lr){4-5} \cmidrule(lr){6-7}
& & & 25th & 75th & Min & Max \\
\midrule
10 & 0.252 & 0.268 & 0.107 & 0.299 & 0.018 & 0.714 \\
30 & 0.707 & 0.795 & 0.580 & 0.853 & 0.304 & 0.911 \\
50 & 0.726 & 0.759 & 0.589 & 0.857 & 0.446 & 0.929 \\
\bottomrule
\end{tabular}
\end{minipage}
\hfill
\begin{minipage}{0.48\textwidth}
\centering
\textbf{Choice Perceptron}\\[3pt]
\begin{tabular}{@{}c cccccc@{}}
\toprule
\textbf{Query} & \textbf{Mean} & \textbf{Median} &
\multicolumn{2}{c}{\textbf{Percentile}} &
\multicolumn{2}{c}{\textbf{Range}} \\
\cmidrule(lr){4-5} \cmidrule(lr){6-7}
& & & 25th & 75th & Min & Max \\
\midrule
10 & 0.162 & 0.116 & 0.071 & 0.228 & 0.000 & 0.679 \\
30 & 0.448 & 0.473 & 0.326 & 0.567 & 0.036 & 0.857 \\
50 & 0.388 & 0.446 & 0.165 & 0.585 & 0.000 & 0.911 \\
\bottomrule
\end{tabular}
\end{minipage}

\end{table*}

\end{document}